# Contrastive Feature Induction for Efficient Structure Learning of Conditional Random Fields

Ni Lao, Jun Zhu
Carnegie Mellon University
5000 Forbes Avenue, Pittsburgh, PA 15213
{nlao,junzhu}@cs.cmu.edu

## Abstract

Structure learning of Conditional Random Fields (CRFs) can be cast into L1-regularized optimization problems, which can be solved by efficient gradient-based optimization methods. However, optimizing a fully linked model may require inference on dense graphs which can be time prohibitive and inaccurate. Gain-based or gradient-based feature selection methods avoid this problem by starting from an empty model and incrementally adding top ranked features to it. However, training time with these incremental methods can be dominated by the cost of evaluating the gain or gradient of candidate features for high-dimensional problems. In this study we propose a fast feature evaluation algorithm called Contrastive Feature Induction (CFI) based on Mean Field Contrastive Divergence ($CD^{MF}$). CFI only evaluates a subset of features which involve variables with high *signals* (deviation from mean) or *errors* (prediction residue). We prove that the gradient of candidate features can be represented solely as a function of signals and errors, and that CFI is an efficient approximation of gradient-based evaluation methods. Experiments on synthetic and real datasets show competitive learning speed and accuracy of CFI on pairwise CRFs, compared to state-of-the-art structure learning methods such as full optimization over all features, and Grafting. More interestingly, CFI is not only faster than other methods but also produces models with higher prediction accuracy by focusing on large prediction errors during induction.

## 1. Introduction

Conditional Random Fields (CRFs) [12] are widely used in applications like computer vision, natural language processing, information extraction, relational learning, computational biology etc. However, as the variety and scale of problems increase, hand-crafting random fields become less applicable. Learning random field structure from data not only provides prediction efficiency and accuracy, but also provides insight about the underlying structure in the domain. For a very long time, the dominant solution to this structure learning problem has been greedy local search [2][14]. Because of the greediness of these procedures, features which become useless in later iterations are not removed from the model, which leads to suboptimal models. Recent work has formulated this structure selection problem as an optimization problem by applying L1-regularization to a fully linked model [1]. However, full optimization requires inference on potentially very dense graphs, which is time prohibitive and inaccurate [13]. Approximate inference methods such as loopy Belief Propagation (BP) [16] are likely to give highly inaccurate estimates of the gradient, leading to poorly learned models [13]. Gain-based or gradient-based feature selection methods avoid this problem by starting from an empty model and incrementally adding top ranked features to it [17][13]. However, for high dimensional problems, the training time of these selection methods can be dominated by the cost of evaluating the gain or gradient of candidate features.

When exhaustively evaluating the gradient of candidate features, much of the calculation is wasted, in two ways. First, for a particular feature *f:A=a,B=b* and a particular training instance, if both *A=a* and *B=b* have ignorable prediction *error* (see precise definition in section 3) then the gradient of *f* on this instance is also ignorable. Second, even if *A=a* has significant error in many instances, but *B=b* is not *informative* (having the same expectation in all the instances), then the gradient of *f* on different instances would cancel each other, under a mild assumption that the errors of *A=a* have zero mean when considering all instances. In other words, instead of enumerating all candidate features for evaluation, it is sufficient only to evaluate a subset of features which involve variables with high signal or error. For example, for a pairwise CRFs of 1000 nodes, if only 10% of the variables are considered for a training instance, then the number of features evaluated is reduced from 500k down to 5k, which reduces time spent on evaluation by a factor of 100.

Based on this insight, we propose a fast feature evaluation algorithm called Contrastive Feature Induction (CFI) based on Mean Field Contrastive Divergence ($CD^{MF}$) [22]. We prove that the gradient of candidate features can be represented solely as function of signals and errors, and, by ignoring small signals and errors, CFI efficiently approximates the gradient-based evaluation methods. We test our method on synthetic data generated from known MRFs and on two real datasets of relation learning problems. We found that CFI is not only faster than other methods, but also produces models with higher prediction accuracy by focusing on large prediction errors during induction.

**Related work**: Elidan et al.'s "ideal parent" algorithm [4] for continuous variable Bayes networks provides a nice principle for detecting potential graph structures. It uses a monotonic function of the cosine similarity between signal vectors and error vectors of variables to filter candidate structures before scoring them with more costly BIC [20]. Our CFI algorithm adapts this idea to categorical variable CRFs, but is different in two aspects. First, "ideal parent" uses signals and errors to estimate the gain of Bayes net log-likelihood, while CFI uses them to estimate the gradient of the Mean Field Contrastive Divergence objective function. Second, "ideal parent" is targeted toward reducing the very costly BIC scoring, and therefore evaluates all possible candidate structures. In contrast, CFI is targeted to avoid evaluating all possible candidates for efficient learning of high dimensional problems. This goal is achieved by ignoring small signals and errors. However, the two algorithms are connected, as they both apply a dot product between error and signal vectors. Furthermore, if the error and signal vectors of "ideal parent" are also sparsified (by ignoring small terms), we would expect similar reductions in training time as CFI. These insights suggest a strong connection between continuous and categorical variable structure learning, and we can think of the two algorithms as one principle instantiated in two different types of graphical models.

The rest of this paper is organized as follows. Section 2 introduces preliminary concepts of CRFs. Section 3 describes the CFI method. Section 4 presents and analyzes experimental results on synthetic and real data, and Section 5 concludes with a summary of contributions.

## 2. Preliminaries

Conditional Random Fields (CRFs [12]) are Markov Random Fields (MRFs) that are globally conditioned on observations. We use upper case letters like *X, Y* for random variables, and lower case letters like *x, y* for variable assignments. Bolded letters are vectors of variables or assignments. Let *G=(V,E)* be an undirected model, over observed variables ***O***, *labeled* variables ***Y***, and *hidden* variables ***H***. Here we consider the general case with hidden variables and it subsumes regular CRFs without hidden variables when ***H*** is empty. Let ***X=(Y,H)*** be the joint of hidden and labeled variables. Then ***X*** has the distribution (1) according to the model. *Feature functions* $f_k(\boldsymbol{x},\boldsymbol{o})$ count how many times a feature fires in the CRFs. $\theta$ is a vector of feature weights. $Z(\theta)$ is called the *partition function*.

$$p(\boldsymbol{x}/\boldsymbol{o};\theta) = \exp(\theta^T \boldsymbol{f}(\boldsymbol{x},\boldsymbol{o}))/Z(\theta)$$

$$\boldsymbol{f}(\boldsymbol{x},\boldsymbol{o}) = [...f_k(\boldsymbol{x},\boldsymbol{o})...]_{k=1..K}^T$$

(1)

For general structure CRFs, inference cannot be performed efficiently with dynamic programming. In this case, $p(\boldsymbol{x}/\boldsymbol{o};\theta)$ can be approximated by methods like belief propagation [15], or mean field approximation [8]. In this study we use mean field approximation, which approximates the true distribution $p(\bm{x}|\bm{o};\theta)$ with a fully factorized distribution $q(x)$ as in Eq. (2) by minimizing their KL divergence. The variational parameters $\mu(X_k=x_k)$ are updated iteratively according to Eq. (3). $nb(f_r)$ is the set of *states* (e.g. $X_k=x_k$) involved in feature $f_r$. $nb(X_k=x_k)$ is the set of features involving state $X_k=x_k$. The expectation of a feature being activated is given by Eq. (4).

$$p(\bm{x}|\bm{o};\theta) \approx q(x) = \prod_k q(X_k = x_k) = \prod_k \mu(X_k = x_k) \quad (2)$$

$$\mu^{new}(X_k = x_k) = \frac{1}{Z_k} \exp\left[\sum_{f_r \in nb(X_k=x_k)} \theta_r \prod_{X_j=x_j \in nb(f_r)\setminus\{X_k=x_k\}} \mu(X_j = x_j)\right] \quad (3)$$

$$\langle f_r \rangle = \prod_{X_j=x_j \in nb(f_r)} \mu(X_j = x_j) \quad (4)$$

Given a set of labeled training data $D=\{(\bm{y}^i,\bm{o}^i)\}$, parameter learning of the CRFs can be formulated as maximizing regularized log-likelihood as in Eq. (5), where $\lambda_1$ controls L1-regularization to help structure selection [1], and $\lambda_2$ controls L2- regularization to prevent over fitting. Differentiating $l(\theta)$ with respect to $\theta$, we get its gradient $\bm{g}$ as in Eq. (7), where $\langle \Box \rangle_p$ is the expectation under the distribution $p$.

$$L(\theta) = \sum_{m=1..M} l_m(\theta) - \lambda_1 |\theta|_1 - \lambda_2 |\theta|_2 / 2 \quad (5)$$

$$l(\theta) = \log p(\bm{y}|\bm{o};\theta) = \log \sum_h p(\bm{y},\bm{h}|\bm{o};\theta) \quad (6)$$

$$\bm{g}(\theta) = \partial l(\theta)/\partial \theta = \langle \bm{f} \rangle_{p(\bm{h},\bm{y}|\bm{o};\theta)} - \langle \bm{f} \rangle_{p(\bm{h}|\bm{o},\bm{y};\theta)} \quad (7)$$

Since $l(\theta)$ and $g(\theta)$ are intractable to compute, we use 1-step Mean Field Contrastive Divergence ($CD^{MF}$) [22] as an approximation. The objective function CD in Eq. (8) is an approximation to a lower bound of $l(\theta)$. Here $q_0$ is the mean-field approximation of $p(\bm{h}|\bm{y},\bm{o};\theta)$. $q_1$ is the approximation of $p(\bm{h},\bm{y}|\bm{o};\theta)$ obtained by applying one mean field update to $q_0$. Generally, $F_t$ is the free energy of the model at $t$-th mean-field update, and $F_0 \geq F \geq F_\infty$. 1-step $CD^{MF}$ utilizes $F_1$ as approximation of $F_\infty$. One main advantage of CD is that it avoids been trapped in possible multimodal distribution of $p(\bm{h},\bm{y}|\bm{o};\theta)$. By starting from $q_0$, $q_1$ always goes to the same mode as $q_0$ [22]. When there are no hidden variables, $q_0$ is just the empirical distribution in training data. When there are hidden variables, inference is needed to estimate $q_0$.

$$l(\theta) \geq \sum_h q(h)\ln\frac{p(y,h;\theta)}{q(h)} = F_\infty - F_0 \approx F_1 - F_0 = KL(q_1 \| p) - KL(q_0 \| p) = CD(\theta) \quad (8)$$

$$KL[q_i \| p] = -\langle \ln p(h,y|o)\rangle_{q_i} - H(q_i) \quad (9)$$

$$\bm{g}(\theta) = \partial CD(\theta)/\partial \theta = \langle \bm{f} \rangle_{q_1} - \langle \bm{f} \rangle_{q_0} \quad (10)$$

$$CD(\theta) = \langle \theta^T \bm{f}\rangle_{q_0} + H(q_0) - \langle \theta^T \bm{f}\rangle_{q_1} - H(q_1) = -\theta^T \bm{g}(\theta) + H(q_0) - H(q_1) \quad (11)$$

We use orthant-wise L-BFGS [1] to tune $\theta$. Update is terminated if the change of both objective and $\|\theta\|_1$ are smaller than a threshold for several consecutive iterations.

## 3. Contrastive Feature Induction (CFI)

Although we define CRFs with hidden variables in the previous section, CFI can be applied to CRFs with or without hidden variables. We just refer to each training sample as $(\bm{x}^i,\bm{o}^i)$ to subsume both cases. Let's first consider *pairwise Markov networks*, where each feature function is an indicator function for a certain assignment to either a pair of nodes or a single node. As we have stated earlier, training time of existing gain-based or gradient-based feature selection methods can be dominated by evaluating the gain or gradient of candidate features for high dimensional problems. Here we present a fast approximation method.

Before diving into the exposition of the algorithm details, let's first define two key concepts. Let

$q_0(X^i=x)$ and $q_1(X^i=x)$ be the estimations of variable $X$ having value $x$ in instance $(\boldsymbol{x}^i,\boldsymbol{o}^i)$ under distribution $q_0$ and $q_1$ as defined in Section 2. We define the error and signal of a state as follows:

**Definition 1**: We define *error* of state $X=x$ in instance $(\boldsymbol{x}^i,\boldsymbol{o}^i)$ be $err(X^i=x)=q_1(X^i=x) - q_0(X^i=x)$.

**Definition 2**: We define *signal* of state $X=x$ in instance $(\boldsymbol{x}^i,\boldsymbol{o}^i)$ be $\varepsilon_t(X^i=x)=q_t(X^i=x) - E_t[X=x]$, where $E_t[X=x]=\sum_j q_t(X^j=x)/M$ is the mean of $q_t(X^i=x)$ in all instances under distribution $q_t$.

We found that much of the calculation in Grafting is wasted, in two ways. First, for a particular feature $f:A=a,B=b$ and a particular training sample $(\boldsymbol{x}^i,\boldsymbol{o}^i)$, if both $A=a$ and $B=b$ have ignorable prediction error then from Eq. (4) we know that $g^i(f)$, the gradient of $f$ on sample $(\boldsymbol{x}^i,\boldsymbol{o}^i)$, is ignorable. Second, even if $A=a$ has significant error in many instances, but $B=b$ is not *informative* (having the same expectation in all the instances) then $\sum_i g^i(f)$ is zero assuming the error of $A=a$ has zero mean when considering all the training samples. This is a reasonable assumption since single node features generally converge quickly. In other words, instead of enumerating all candidate features for evaluation, it is sufficient to only evaluate a subset which involves variables with high signal or error. For example, for the pairwise CRFs of 1000 nodes, if only 10% of the variables are considered for a sample, then the number of features evaluated is reduced from 500k down to 5k, which reduces training time by a factor of 100.

More formally, we prove that CFI is an efficient approximation to the gradient based evaluation methods.

**Theorem 3 (Gradient Decomposition Theorem: GDT)**: The gradient in Eq. (10) can be represented purely as a function $F(\cdot)$ of signals and errors

$$g(\theta_{A=a,B=b}) = \sum_i F(\varepsilon_0(A^i=a), \varepsilon_0(B^i=b), err(A^i=a), err(B^i=b))$$

**Proof:** As in Eq. (12) and (13), $g$ can be decomposed into signals, errors and means. Assuming weights of unary features have already converged, we have $g(\theta_{X=x})/M=E_1[X=x]-E_0[X=x]=0$ (or $E_1[X=x]=E_0[X=x]=E[X=x]$, regularization is ignored here). Then the last two terms in Eq. (13) will be canceled in Eq. (12) with other samples

$$g(\theta_{A=a,B=b}) = \sum_i g^i(\theta_{A=a,B=b}) \tag{12}$$

$$\begin{aligned}
g^i(\theta_{A=a,B=b}) &= q_1(A^i=a)q_1(B^i=b) - q_0(A^i=a)q_0(B^i=b) \\
&= err(A^i=a)\varepsilon_0(B^i=b) + err(B^i=b)\varepsilon_0(A^i=a) + err(A^i=a)err(B^i=b) \\
&\quad + err(A^i=a)E_0[B^i=b] + err(B^i=b)E_0[A^i=a] \\
&= [\varepsilon_1(A^i=a) + \varepsilon_0(A^i=a)]*err(B^i=b)/2 + [\varepsilon_1(B^i=b) + \varepsilon_0(B^i=b)]*err(A^i=a)/2 \\
&\quad + err(A^i=a)E_0[B^i=b] + err(B^i=b)E_0[A^i=a]
\end{aligned} \tag{13}$$

If we further ignore small errors and signals with magnitudes smaller than thresholds $t_{err}$ and $t_{sig}$ in Eq. (13), we can effectively ignore many terms in Eq. (12). At each training iteration, the Contrastive Gradient Approximation algorithm (as outlined in Algorithm 1) estimates a sparse vector $(h_{A=a,B=b})$ as an approximation of $(g(\theta_{A=a,B=b}))$, and adds top $J$ features to the model. We call $J$ the *batch size*.

---

**Algorithm 1: Contrastive Gradient Approximation**
**Input**: $\varepsilon_0$, $\varepsilon_1$, $err$ of all states of all training samples
**Output**: sparse vector $(h_{A=a,B=b})$
Set $h_{A=a,B=b}=0$, for any pair of states $A=a, B=b$
For each sample $(\boldsymbol{x}^i,\boldsymbol{o}^i)$
  Find the set of states $S_{err}(i)$ with $|err| > t_{err}$,
  Find the set of states $S_{sig}(i)$ with $|\varepsilon_0+\varepsilon_1|/2 > t_{sig}$.
  For any pair $A^i=a \in S_{sig}(i)$, $B^i=b \in S_{err}(i)$,
    Set $h_{A=a,B=b}+=[\varepsilon_0(A^i=a)+ \varepsilon_1(A^i=a)]* err(B^i=b)/2$.

---

To fulfill the assumption that unary features have already converged, we can split the training into two stages: the first stage adds all unary features to the model and optimizes until convergence; the second stage incrementally induces pairwise features during optimization. In our experience, however, merging the two stages into one does not affect the accuracy of the learned model. So we start inducing binary features right after adding all the unary features, and save the iterations (usually 5) spent on waiting for unary features to converge.

The CFI algorithm shares the same principle suggested by the "ideal parent" algorithm [4] — if one variable has high correlation to another variable's prediction error, a link should be formed between them. If we concatenate errors and signals of a state in all samples into two vectors, then the approximated feature gradients given by CFI are exactly the dot products between sparse error vectors and signal vectors. This suggests the connection between CFI and "ideal parent" which also involves dot product between error vectors and signal vectors. Furthermore, if the error and signal vectors of "ideal parent" are also sparsified (by ignoring small terms), we would expect similar reductions in training time as CFI on high dimensional problems. We can think the two algorithms as one principle instantiated in two different types of graphical models.

To go beyond pairwise networks, similar techniques can be used to decompose the gradient of higher order features $f:X_1=x_1,...,X_K=x_K$ as in Eq. (14).

$$\begin{aligned} g^i(\theta_{X_1=x_1,...X_K=x_K}) &= \prod_k q_1(X_k^i = x_k) - \prod_k q_0(X_k^i = x_k) \\ &= \prod_k \{q_0(X_k^i = x_k) + err[X_k^i = x_k]\} - \prod_k q_0(X_k^i = x_k) \end{aligned} \quad (14)$$

Although GDT hinges on the independence of variables (mean field estimation) when calculating the gradient of features, it does not restrict the type of inference used to optimize the model. One can surely induce features with CFI while optimizing the model with loopy BP. However, it remains an interesting research question whether similar decomposition results can be found for other types of objective functions that do not have as strong independence assumption as mean field.

## 4. Experiment

In this study, we report empirical results for pairwise networks by comparing CFI with state-of-the-art structure learning methods including full optimization over all features (Full-L1) [1], and Grafting [17]. The original Grafting method optimizes the model until convergence after adding each batch of new features. Empirically we found it is more efficient to update only one iteration after each batch of features are added, and the quality of the learnt models is not sacrificed.

Following the methodology of Kok and Domingos [10], we situate our experiment in prediction tasks. Ten-fold cross validation is performed as follows. We first mark a randomly selected 10% labels in the data as hidden during training, and then evaluate on these labels during testing, conditioned on the observed training labels. Performances of the systems are measured by training time, prediction error rate, average Conditional Log-Likelihood (CLL), and Area Under precision-recall Curve (AUC) [10]. CLL is calculated by $\sum_i \ln q_0(h^i/y^i)$, where $h^i$ are the marked 10% of the labels, and $y^i$ are the remaining 90% of the labels. AUC is calculated in three steps. We first sort all states of the testing variables in decreasing order of $q_0(H^i=h^i/y^i)$, then treat all the true states as relevant and calculate precision and recall at each position. Finally, we integrate the area under the precision/recall curve. In all experiments we fix $\lambda_2=1$, and $\lambda_1$ is grossly tuned for Grafting. We found that within the range of 30-100, batch size $J$ does not affect the performance of Grafting and CFI much, so we set the batch size to 50 in all experiments. All algorithms are implemented in Java 6.0.

### 4.1. Synthetic Data

Following the method described by Lee et al. [13] we generated synthetic data through Gibbs sampling (10000 iteration burn-in, 1000 iteration thinning) on synthetic networks. A network structure with $N$ nodes was generated by treating each possible edge as a Bernoulli random variable and sampling the edges. We chose the parameter of Bernoulli distribution so that each node had $K$ neighbors on average. Weights of the chosen edges are drawn from an even distribution within [-5, 5]. We fix $\lambda_1=2$, $M=200$ (number of sample), $k=5$, and vary $N$. We also compare to TrueGraph, for which features in the true models are optimized and no feature induction is applied.

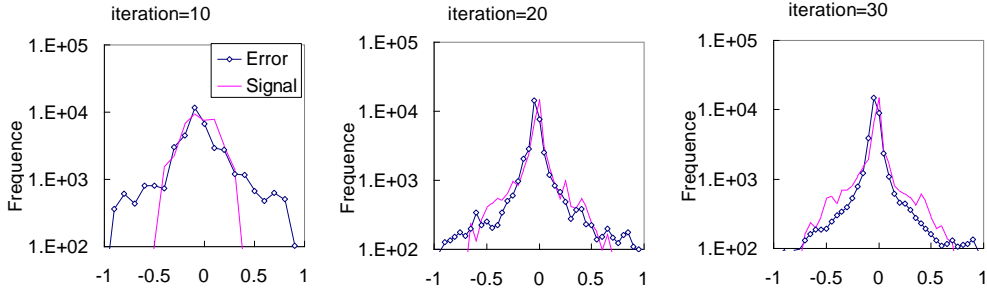

Figure 1  Histogram of signal and error using CFI induction with $N$=200, $\lambda_1$=2. Y-axis is the number of states across all instances with corresponding error/signal on X-axis.

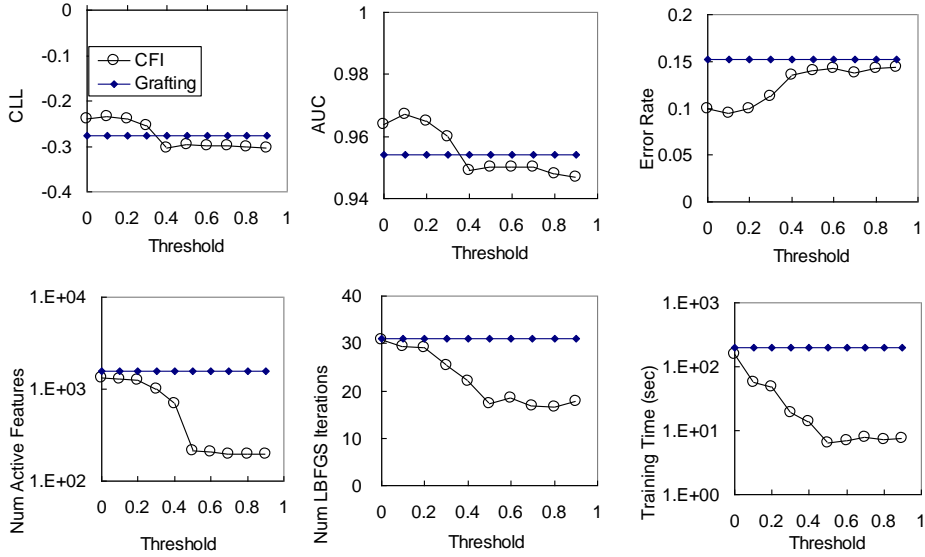

Figure 2  Results of CFI method by using different threshold $t_{sig}=t_{err}$ with $N$ =200, $\lambda_1$=2.

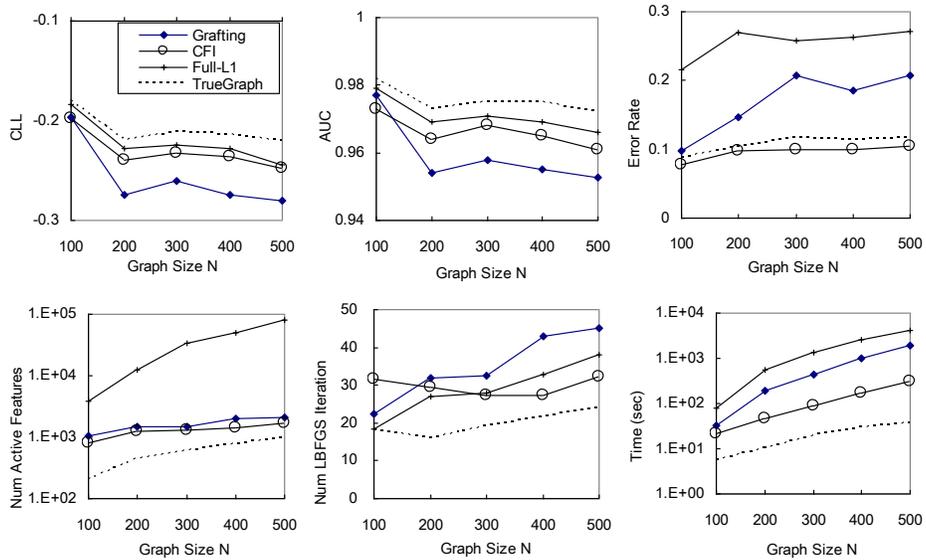

Figure 3  The performance of different methods on different network size $N$, with $\lambda_1$=2.

To check the distribution of signals and errors we draw their histograms in Figure 1. We can see that since signals and errors mostly concentrate around 0, substantial saving of computation can be expected from ignoring small terms. In this study we set $t_{err}$ and $t_{eps}$ to have the same value. Figure 2 shows relation between training time and the threshold. We can see that as $t_{sig}$ and $t_{err}$ gets larger, CFI is increasingly faster. When $t_{err}=t_{eps}=0$, CFI falls back to similar training time as Grafting. There is a sharp drop in number of features when $t_{err}=t_{sig}$ passes 0.5, because most signals and errors have magnitude smaller than 0.5. More interestingly, when thresholds are not too large (<0.4), CFI leads to better prediction than Grafting. In the rest of this study, we fix $t_{err}=t_{sig}=0.2$.

Figure 3 compares different methods on different network size. We can see that among the three methods, full-L1 is the slowest, because it lets all features participate in inference, but it also has the best CLL and AUC. CFI is not only the fastest but also has the smallest error rate. This shows that CFI has an effect similar to that observed in large margin approaches, where the focus is on reducing large prediction errors which potentially leads to flip of decisions. CFI even has a smaller error rate than the true graph, because features not in the true graph are also useful in prediction. Note that if we use a more costly inference method like loopy BP rather than mean field, then the time difference between full-L1 and the other two methods would be larger. CFI produces slightly less number of active features (features with non-zero weights) than Grafting, even though they introduce almost the same number of features in total.

### 4.2. Relational Learning Data

There has been a surge of interests in Statistical Relational Learning (SRL [5]) driven by applications like collective classification, information extraction and integration, bioinformatics, et al. Among many proposed SRL models, Template-Based Relation Markov Random Fields (TR-MRFs [7]) is a powerful model. It does not have the acyclic constraint as Probabilistic Relational Model (PRM) (which is based on Bayes network), nor does it confine itself to binary random variables as Markov Logic Network (MLN). The binary representation of MLN causes deterministic dependencies when modeling multi-category concepts, and requires special treatment like slice-sampling during inference to help find separate modes of distribution [19]. However, there is no previous work on learning structure of TR-MRFs. In this section, we compare different structure learning methods for TR-MRFs, and compare their performance to other state of the art relational learning models.

TR-MRFs define a random field on a set of *entities*. Each entity can be seen as a training instance, and has a set of *attributes* (variables). However, the entities are not independent; features can be defined between attributes of different entities through the concept of *templates*. Therefore, all the training samples are connected into a single random field and are collectively called a *mega-sample*. For formal definitions, please refer to Jaimovich et al. [7]; we leave out implementation details here due to the limitation of space.

Note that instead of using lifted belief propagation for inference [7], we use generalized mean field approximation [23], which is fast and easier to implement. We treat each entity as a sub-graph. The joint distribution of all entities is estimated by doing mean field update one entity at a time with the variational parameters of its neighbors fixed. The joint distribution converges after iterating through all entities for several rounds. Within each round, the entities are visited with a fixed order, and iteration terminates when maximum change of expectation in a round is smaller than $\varepsilon=0.1$.

We used two relational datasets that have relatively large number of attributes (available at http://alchemy.cs.washington.edu [10]). The **Animal** dataset contains a set of animals and their attributes. It consists exclusively of unary predicates of the form f($a$), where f is a feature and $a$ is an animal (e.g., Swims(Dolphin)). There are 50 animals, and 85 attributes. This is a simple propositional dataset with no relational structure, but is useful as a "base case" for comparison. There are 3,655 candidate features in its pairwise random field. The **Nation** dataset contains attributes of nations and relations among them. The binary predicates are of the form r($n$, $n0$), where $n$, $n0$ are nations, and r is a relation between them (e.g., *ExportsTo*, *GivesEconomicAidTo*). The unary predicates are of the form f($n$), where $n$ is a nation and f is a feature (e.g., *Communist*, *Monarchy*). There are 14 nations, 56 relations and 111 attributes. There are 20,244 candidate features in its pairwise random field.

We compare TR-MRF with the results of state-of-the-art relational learning algorithms reported in [10]. The **MLN structure learning algorithm** (MSL [11]) creates candidate clauses by adding literals to the current clauses. The weight of each candidate clause is learned by optimizing a weighted pseudo-log-likelihood (WPLL) measure, and the best one is added to the MLN. There are two relational learning algorithms that induce hidden variables. **Infinite Relational Model** (IRM [9]) simultaneously clusters objects and relations. It defines a generative model for the predicates and cluster assignments. It uses a Chinese restaurant process prior (CRP [18]) on the cluster assignments to automatically control number of clusters. **Multiple Relational Clusterings** (MRC [10]) is based on MLN. Similar to IRM, it automatically invents predicates by clustering objects, attributes and relations. Moreover, MRC can learn multiple clusterings, rather than just one.

Table 1 compares performance of different methods on the two data sets, where the result of MSL/IRM/MRC are directly cited from [10]. Kok & Domingos [10] did not report error rates. We can see that when the number of candidate features is small (3,655 for Animal data), selectively adding features (Grafting and CFI) does not benefit learning speed or prediction accuracy. However, when the problem dimension is large (20,244 for Nation data), avoiding training on a dense graph becomes more important. For CLL and AUC, CFI performs significantly better than Grafting, which is significantly better than Full-L1. For error rate, CFI and Grafting are not significantly different, but both are significantly better than Full-L1. This indicates that adding top-ranked features instead of all features to a model is beneficial for prediction tasks. Although MLN is more flexible in defining complex features (clauses) than pairwise fields, it is very challenging to search through the large space of possible structures, which is evidenced by the much larger training time. As shown in Figure 4, although errors mostly concentrate around 0, signals have a two mode distribution on this data. Therefore, the saving of time is mainly from considering less number of error terms.

Table 1 Comparison to state of the art relational learning methods. The numbers in parentheses are standard deviations.

|  | **Animal**, $\lambda_1=0.5$ | | | | **Nation**, $\lambda_1=2$ | | | |
|---|---|---|---|---|---|---|---|---|
|  | **AUC** | **CLL** | **Err** | **Time** | **AUC** | **CLL** | **Err** | **Time** |
| **MSL** | 68(4) | -0.54(0.04) | | <24hour | 77(4) | -0.33(0.04) | | 24hour |
| **IRM** | 79(8) | -0.43(0.06) | | 10hour | 75(3) | -0.32(0.02) | | 10hour |
| **MRC** | 80(4) | -0.43(0.04) | | 10hour | 75(3) | -0.31(0.02) | | 10hour |
| **Full-L1** | 92.8(1) | -0.337(0.02) | 14.3(1) | 0.17(0.01)min | 88.8(1) | -0.41(0.02) | 16.0(1) | 6.5(0.3) min |
| **Grafting** | 91.6(1) | -0.361(0.03) | 14.9(1) | 0.19(0.01)min | 90.2(1) | -0.37(0.04) | 11.7(2) | 4.2(0.9) min |
| **CFI** | 92.1(1) | -0.355(0.04) | 14.9(1) | 0.16(0.01)min | 93.6(1) | -0.30(0.02) | 11.2(1) | 0.54(0.1) min |

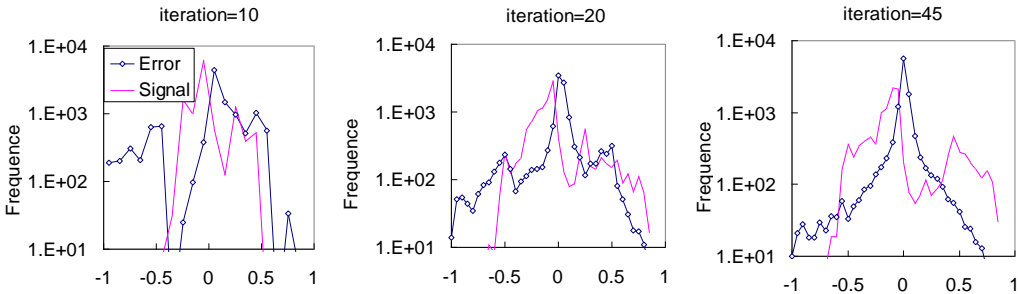

Figure 4 Histogram of signals and errors on Nation data with CFI induction and $\lambda_1=2$.

## 5. Conclusion

In this study we propose a fast feature evaluation algorithm called Contrastive Feature Induction (CFI). It evaluates a subset of features which involve variables with high *signal* (deviation from mean) or *error* (prediction residues). We prove that CFI is an efficient approximation of the gradient-based evaluation methods. Experiments on synthetic and real datasets show that CFI is not only faster than Grafting and full optimization of all the features, but also produces models of higher prediction accuracy by focusing on large prediction errors.